\documentclass{article}
\usepackage[T1]{fontenc}

\usepackage{amsfonts} 
\usepackage{amsmath}  
\usepackage{xspace}
\usepackage{subcaption}

\usepackage{tikz}
\usetikzlibrary{arrows.meta,calc,positioning}
\definecolor{InfColor}{HTML}{D97925}
\definecolor{FTFwdColor}{HTML}{8CBBD9}
\definecolor{BwdColor}{HTML}{3F6FA5}
\definecolor{BurstColor}{HTML}{B73A2D}
\definecolor{BurstFill}{HTML}{F6E5E2}
\definecolor{MutedColor}{HTML}{626262}

\newcommand{\name}{{LayerLoom}\xspace}

\usepackage{microtype}
\usepackage{graphicx}
\usepackage{booktabs} 

\usepackage{hyperref}

\usepackage{algorithm}
\usepackage{algorithmic}

\usepackage[accepted]{mlsys2025}

\usepackage{enumitem}
\setlist[itemize]{topsep=2pt,partopsep=0pt,
  itemsep=0pt,parsep=0pt,leftmargin=*}

\mlsystitlerunning{Inference First, Training Underneath: Layer-Granular LoRA Co-Serving with \name}

\begin{document}

\twocolumn[
\mlsystitle{Inference First, Training Underneath: \\Layer-Granular LoRA Co-Serving with \name}



\mlsyssetsymbol{equal}{*}

\begin{mlsysauthorlist}
\mlsysauthor{Zongpu Zhang}{ind}
\mlsysauthor{Yishu Yin}{thu}
\mlsysauthor{Xuehai Qian}{thu}
\end{mlsysauthorlist}

\mlsysaffiliation{ind}{Independent Researcher}
\mlsysaffiliation{thu}{Tsinghua University, China}


\mlsyskeywords{Machine Learning, MLSys}

\vskip 0.3in

\begin{abstract}
Serving and fine-tuning a large language model on the same GPUs can turn the idle capacity between inference bursts into training progress, and low-rank adaptation (LoRA) makes this co-location cheap because both workloads read a single frozen copy of the base weights.
The difficulty lies in the fine-tuning backward pass, which has no counterpart in inference.
Current co-serving systems schedule backward work per serving iteration: each iteration carries a fixed backward budget, and requests that arrive in the meantime wait for it to finish, so fine-tuning dictates inference latency whenever load spikes.
We observe that the backward pass of a LoRA microbatch decomposes into per-layer tasks over the same weights that the inference forward pass visits, which allows backward work to be scheduled at layer rather than iteration granularity.
Building on this observation, \name attaches same-layer backward tasks to the inference forward pass and, at each layer, runs only as many of them as the latency slack of in-flight requests permits.
To keep a ready task at every layer, \name maintains a wavefront of fine-tuning microbatches staggered across layers, and it bounds the wavefront's activation memory by offloading microbatch state to host memory and prefetching it ahead of the layer where the scheduler plans to use it.
A fused kernel serves the paired forward and backward GEMMs from a single read of each base weight.
Because all backward work of an iteration finishes within its forward pass, new requests join the next iteration as in continuous batching.
On an RTX 4090 GPU running Llama-3.1-8B under a bursty production trace, \name raises joint SLO attainment by 13.5--21.8 points over prior works with fine-tuning throughput that is comparable or up to 12\% higher, and it reaches the same training losses.
\end{abstract}

]



\begingroup
\renewcommand{\thefootnote}{}
\footnotetext{\textsuperscript{1}Independent Researcher.
  \textsuperscript{2}Tsinghua University, China.}
\endgroup

\section{Introduction}
\label{sec:introduction}

LLM platforms increasingly serve a model online while also fine-tuning it,
e.g., to adapt it to new domains or users. The two workloads have opposite
performance objectives. Online inference must meet a latency target for every
request, i.e., a time-to-first-token (TTFT) and a time-per-output-token (TPOT)
target, whereas fine-tuning prefers sustained throughput. Since inference
demand is bursty, serving GPUs are typically provisioned for demand peaks and
remain underutilized between bursts~\cite{wang2025burstgpt}. Running
fine-tuning on the same GPUs could turn this idle capacity into training
progress, provided that fine-tuning yields the GPU as soon as inference needs
it.

Low-rank adaptation (LoRA)~\cite{hu2022lora} makes such co-serving
particularly attractive. Since LoRA freezes the pretrained base weights and
trains only small adapters, inference and fine-tuning can share one resident
copy of the base weights, and their forward passes apply the same operators to
different tokens. Recent co-serving
systems~\cite{he2025resource,zhang2026loquetier,oliaro2026flexllm} exploit
this sharing by merging fine-tuning tokens into the inference batch of each
serving iteration.

The fine-tuning backward pass, however, has no counterpart in inference, and
existing systems handle it at the granularity of a serving iteration. They
attach a fixed amount of backward work to each iteration, e.g., after the
merged forward pass or on a separate CUDA stream in
FlexLLM~\cite{oliaro2026flexllm}, and form the next batch only after this work
completes. A request that arrives during an iteration therefore waits for the
attached backward work regardless of its latency target, which makes such
iteration-level co-serving \emph{fine-tuning-prioritized} in effect. Under
bursty production traces, it lowers the joint SLO attainment of FlexLLM from
at least 99\% under inference alone to 1.5--19.0\%, and attaching less
backward work to each iteration restores responsiveness only by cutting
fine-tuning throughput by up to 28$\times$.

We instead observe that the backward work does not have to be decided once
per iteration. An inference iteration traverses the transformer layers in
order, and the backward pass of a LoRA microbatch is itself a sequence of
per-layer tasks on the same base weights. If backward tasks are attached to
the inference forward pass \emph{one layer at a time}, the runtime can decide
at every layer how much backward work to run, and no backward work remains
when the forward pass ends, so a newly arrived request can join the next
iteration just as in continuous batching~\cite{yu2022orca}.

To this end, we present \name, an \emph{inference-prioritized} co-serving
runtime built on three key ideas. First, a forward-anchored backward pipeline
pairs the inference forward pass of each layer with backward tasks of the same
layer. Since a backward pass runs from the last layer to the first, \name
keeps a \emph{wavefront} of microbatches whose next tasks lie at different
layers, so that every layer finds a ready backward task. Second, an SLO-aware
scheduler decides, layer by layer, how many of these tasks to run within the
latency headroom of active requests. Third, since the paired inference forward
and fine-tuning backward read the same base weight, a shared-weight kernel
reads this weight only once for both GEMMs.

Turning these ideas into a working system raises two challenges. First, each
microbatch in the wavefront retains the activations of all layers it has not
yet processed, so a full wavefront over the 32 layers of Llama-3.1-8B retains
16.5$\times$ the activations of a single microbatch, which competes with the
KV cache for GPU memory. \name addresses this challenge with scheduler-guided
offloading, which moves each microbatch's state to host memory and prefetches
it ahead of the layer at which the scheduler plans to run it. Second, the two
fused GEMMs reduce along opposite dimensions of the base weight, so any
mapping of the weight onto GPU thread blocks must reduce one of the outputs
across blocks. \name reduces this cost by completing the wide backward output
within each thread block and reducing only the narrow inference output across
blocks.

We implement \name on FlexLLM and evaluate it on an RTX 4090 GPU with
Llama-3.1-8B under a bursty production trace at two loads. Compared with
FlexLLM's co-serving mode, \name achieves 21.8 and 13.5 points higher joint
SLO attainment at the two loads, while training 12\% more tokens per second at
the lower load and a comparable number at the higher one. Offloading lets the
wavefront hold twice as many microbatches as no-offloading allows, and the
fused kernel runs 1.21--1.58$\times$ faster than matched unfused kernels.
\name also produces the same training losses as FlexLLM.

In summary, we make the following contributions.
\begin{itemize}
  \item We examine existing co-serving runtimes and observe that they commit
    to backward work once per iteration, which makes inference wait for
    fine-tuning under bursty demand (\S\ref{sec:background}).
  \item We design \name, a forward-anchored backward pipeline with a backward
    wavefront and an SLO-aware scheduler (\S\ref{sec:pipeline}).
  \item We address the two resulting challenges with scheduler-guided
    offloading (\S\ref{subsec:design-offload}) and a shared-weight kernel
    (\S\ref{subsec:design-kern}).
  \item We implement a prototype of \name and evaluate it on a bursty
    production trace, showing 13.5--21.8 points higher SLO attainment with
    comparable or up to 12\% higher fine-tuning throughput than a
    state-of-the-art co-serving runtime (\S\ref{sec:eval}).
\end{itemize}

\section{Background and Motivation}
\label{sec:background}

\begin{figure*}[t]
\centering
\begin{minipage}[b]{.24\textwidth}
\centering
\includegraphics[width=\linewidth]{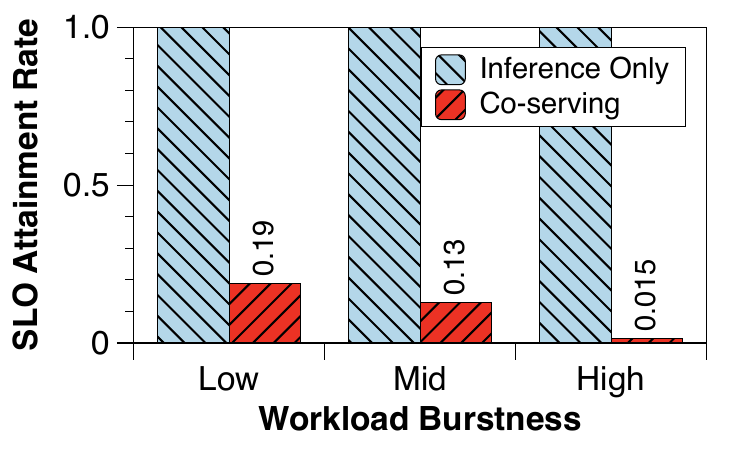}
\subcaption{\centering Joint SLO attainment}
\label{fig:exp-mot-1a}
\end{minipage}\hfill
\begin{minipage}[b]{.24\textwidth}
\centering
\includegraphics[width=\linewidth]{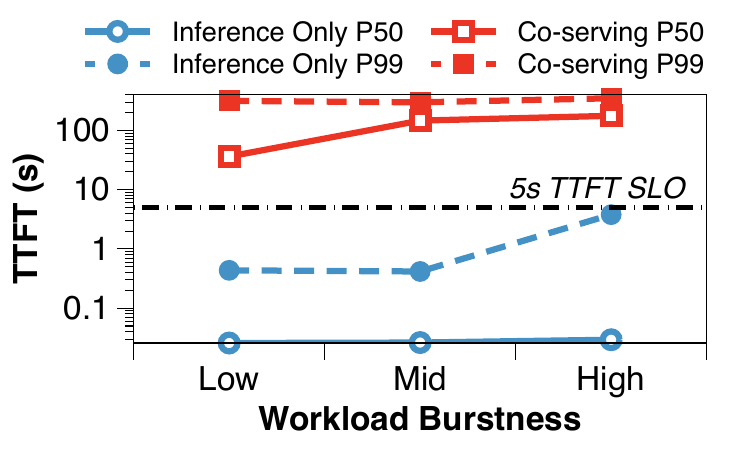}
\subcaption{\centering TTFT (log scale)}
\label{fig:exp-mot-1b}
\end{minipage}\hfill
\begin{minipage}[b]{.24\textwidth}
\centering
\includegraphics[width=\linewidth]{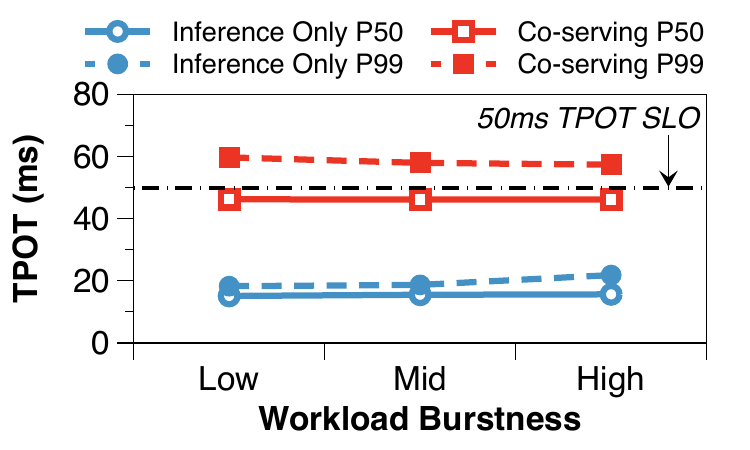}
\subcaption{\centering TPOT}
\label{fig:exp-mot-1c}
\end{minipage}\hfill
\begin{minipage}[b]{.24\textwidth}
\centering
\includegraphics[width=\linewidth]{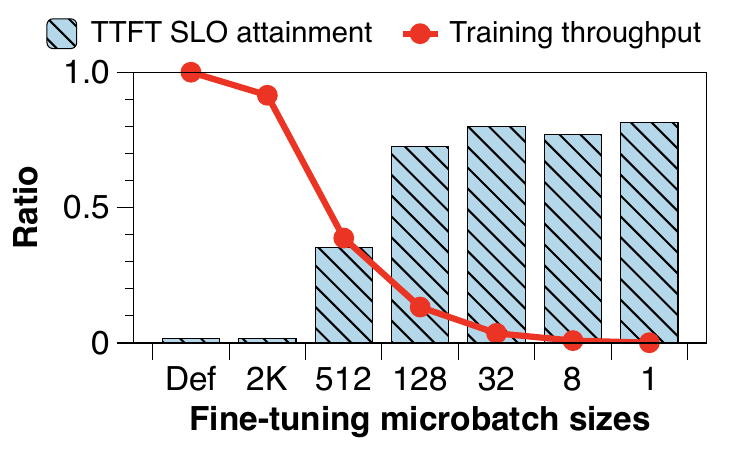}
\subcaption{\centering Capping the microbatch}
\label{fig:motivation_microbatch_tradeoff}
\end{minipage}
\caption{FlexLLM co-serving on an A800 GPU. (a)--(c) Joint SLO attainment,
TTFT, and TPOT under the low-, median-, and high-burstiness traces. (d) Fine-tuning throughput relative to the uncapped
default (line), and TTFT
SLO attainment (bars), when the fine-tuning microbatch is capped at various sizes, under the high
burstiness trace. "Def" refers to FlexLLM's uncapped default microbatches, with a median of
3308 tokens.}
\label{fig:exp-mot-1}
\vspace{-2mm}
\end{figure*}

\subsection{LLM Inference and LoRA Fine-Tuning}
\label{sec:background_lora}

Online LLM serving processes each request in two phases. The \emph{prefill}
phase encodes the prompt into the key-value (KV) cache, and the \emph{decode}
phase then generates one output token at a time from the cached state.
Since the engine can form a new batch at every prefill or decode iteration,
continuous batching~\cite{yu2022orca} allows
newly arrived request to join the batch as soon as the current iteration
finishes.
The latency of a request is accordingly characterized by its time-to-first-token
(TTFT), which includes queueing and prefill, and its time-per-output-token
(TPOT) during decode.

Low-rank adaptation (LoRA)~\cite{hu2022lora} fine-tunes an LLM by freezing its
pretrained base weights and training small low-rank adapters attached to selected
linear layers. Because the base weights are shared, many adapted variants of an LLM can be
served or fine-tuned on one resident copy of its base
weights~\cite{chen2024punica,sheng2024slora,wu2024dlora,xue2026muxtune}.
Consider a LoRA linear layer with frozen base weight
$W \in \mathbb{R}^{d_{\mathrm{out}} \times d_{\mathrm{in}}}$, adapters
$A \in \mathbb{R}^{r \times d_{\mathrm{in}}}$ and
$B \in \mathbb{R}^{d_{\mathrm{out}} \times r}$ with
$r \ll d_{\mathrm{in}},d_{\mathrm{out}}$, and scaling factor $\gamma$. Given an
input $X\in\mathbb{R}^{d_{\mathrm{in}}\times n}$, the layer computes
\begin{equation}
    Y = WX + \gamma B(AX),
    \label{eq:lora_forward}
\end{equation}
and given the output gradient $G=\nabla_Y\mathcal{L}$, backpropagation computes
\begin{equation}
\begin{aligned}
\nabla_X\mathcal{L}&=W^{\mathsf T}G+\gamma A^{\mathsf T}(B^{\mathsf T}G),\\
\nabla_B\mathcal{L}&=\gamma G(AX)^{\mathsf T},\quad
\nabla_A\mathcal{L}=\gamma(B^{\mathsf T}G)X^{\mathsf T}.
\end{aligned}
\label{eq:lora_grad_x}
\end{equation}
Freezing $W$ removes $\nabla_W\mathcal{L}$ and its optimizer state, but
propagating the gradient to earlier layers still requires the dense product
$W^{\mathsf T}G$ in $\nabla_X\mathcal{L}$. The base weight is therefore used both by the
inference and fine-tuning forward passes, which compute $WX$ for different
tokens, and by the fine-tuning backward pass, which computes
$W^{\mathsf T}G$.

\subsection{Co-Serving Inference and Fine-Tuning}
\label{subsec:flexllm}

Since inference GPUs are typically provisioned for demand peaks, they sit partly idle
between bursts, and co-serving can use this idle capacity to fine-tune the same
model. Unlike general GPU-sharing systems, which colocate independent jobs by
time slicing or spatial
partitioning~\cite{strati2024orion,zhang2025improving,wang2025colocating},
LoRA co-serving runtimes such as
FlexLLM~\cite{oliaro2026flexllm} additionally merges the fine-tuning forward 
tokens into the inference batch of each iteration.

The LoRA fine-tuning backward pass, however, has no counterpart in inference and
must be scheduled on its own. Existing runtimes attach a fixed amount of
backward work, e.g., the backward of a few layers of one microbatch, to each
serving iteration. FlexLLM offers three ways to execute this work.
(1) \emph{Co-serving} runs it inside the inference iteration, right after the
merged forward pass, (2) \emph{spatial sharing} runs it on a separate CUDA
stream that overlaps the merged forward pass, while (3) \emph{temporal sharing} 
alternates inference iterations with standalone fine-tuning forward and backward passes.
In all three cases, the runtime forms the next inference batch only after this backward work completes.

Such \emph{iteration-level co-serving} keeps scheduling simple, since each
iteration is a self-contained unit that the runtime can launch.
However, it also means that the runtime commits to the backward work
when it forms an iteration and cannot revise this commitment when new
requests arrive while the iteration runs.

\begin{figure*}[t]
\centering
\includegraphics[width=\textwidth,trim={0bp 43bp 0bp 29bp},clip]{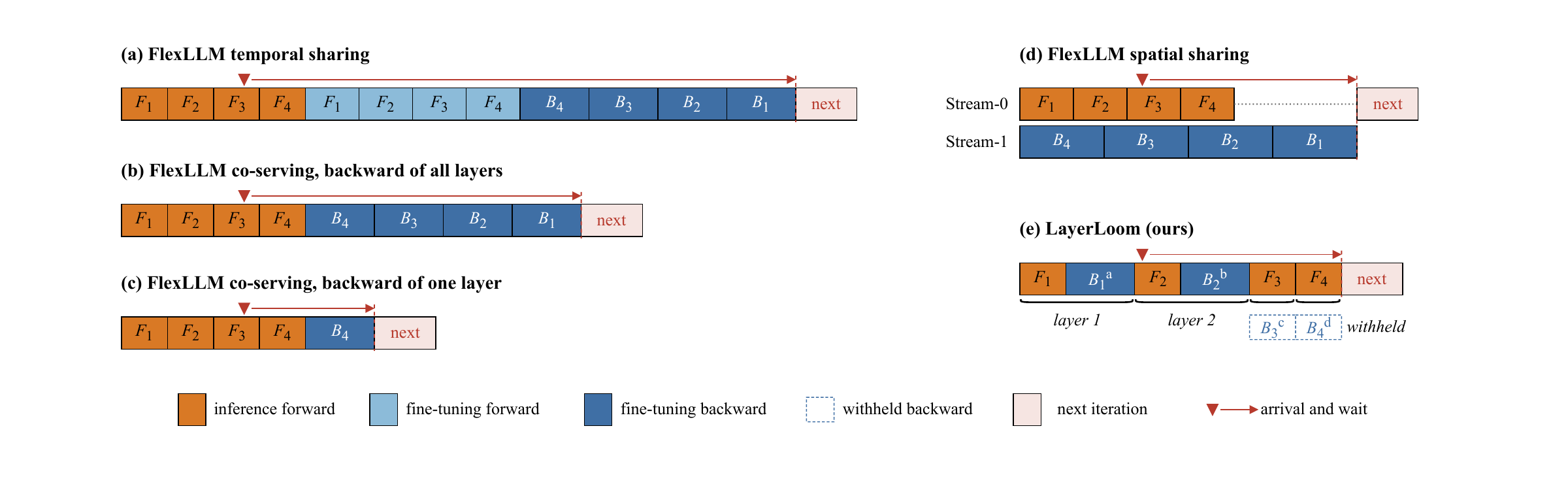}
\caption{One co-serving iteration of a four-layer model, with a request
arriving at the same time in every panel. $F_l$ and $B_l$ are the forward and
backward of layer $l$, and superscripts in (e) denote different
microbatches.}
\label{fig:pipeline-modes}
\vspace{-2mm}
\end{figure*}

\subsection{Motivation}
\label{subsec:motivation}

\noindent\textbf{Setup.}
To examine whether iteration-level co-serving keeps inference within
its latency SLOs when demand is bursty, we run the default co-serving mode of FlexLLM~\cite{oliaro2026flexllm}
on one A800-SXM4-80GB GPU with Llama-3.1-8B model and rank-16 LoRA on the
MLP down-projection, following the FlexLLM paper. Fine-tuning samples are
drawn from Sky-T1 with up to 8K tokens, and inference requests take their
prompt and decode lengths from ShareGPT~\cite{vicuna2023}. To isolate the
effect of burstiness, we cut BurstGPT~\cite{wang2025burstgpt} into segments of
4K consecutive requests, normalize them to a common duration, and pick the
segments at the 10th, 50th, and 90th percentiles of the coefficient of
variation of per-second arrivals as the low, median, and high-burstiness
traces. They share the same sequence of prompt and output lengths and differ only in request timing.

Following FlexLLM's setting for
8B models, a request meets its SLO only if its TTFT is at most 5~s and its
TPOT is at most 50~ms. We replay each trace three times at 2.75 requests/s, the
highest rate at which inference alone attains at least 99\% SLO attainment on
all three traces, so that SLO violations reflect co-serving rather than
overload.

\noindent\textbf{Observation 1: Iteration-level co-serving violates inference SLOs under
bursts.}
As shown in Fig.~\ref{fig:exp-mot-1a}, co-serving lowers the joint (TTFT and TPOT) SLO
attainment from at least 99\% under inference alone to 18.96\%, 12.89\%, and
1.51\% under low, median, and high burstiness, respectively.
Figs.~\ref{fig:exp-mot-1b} and~\ref{fig:exp-mot-1c} further show that the degradation comes almost
entirely from the first token. The median TTFT rises from 36.2~s under low
burstiness to 145.9~s and 175.8~s under median and high burstiness, and the
P99 TTFT exceeds 296~s in every case, whereas the median TPOT stays near
46~ms. This is because a request that arrives during a burst waits behind the
backward work of every iteration ahead of it, and the more requests arrive
together, the longer this queue grows.

\noindent\textbf{Observation 2: Attaching less backward work to each iteration sacrifices
fine-tuning throughput.}
A natural remedy within iteration-level co-serving is to attach less backward
work to each iteration, so that every iteration, and hence every wait, becomes
shorter. Preempting backward work when a request
arrives~\cite{bai2020pipeswitch,han2022microsecond} is an alternative, but the
request would then still wait for the active unit or pay to save and restore
its partial state. To quantify this trade-off, we cap the fine-tuning
microbatch on the high-burstiness trace at 2048, 512, 128, 32, 8, and 1
tokens, whereas FlexLLM's uncapped default uses microbatches with a median of
3308 tokens.

We observe that smaller caps can raise the TTFT SLO attainment
(Fig.~\ref{fig:motivation_microbatch_tradeoff}), from 1.59\% without a cap to
35.29\% at 512 tokens and 72.51\% at 128 tokens, yet even a one-token cap
reaches only 81.40\%, compared with 99.89\% for inference alone. Meanwhile,
fine-tuning throughput drops much faster,
as the 512- and 128-token caps retain only 38.76\% and 13.28\% of the default 2{,}392.7 tokens/s and the
smallest caps retain at most 3.58\%.

\noindent\textbf{Key insight.}
The above results suggest that the amount of backward work per iteration
determines both (a) how long a new request waits and (b) how efficiently the
backward GEMMs run.
A large amount delays every inference request that arrives during a burst
until the work completes, whereas a small amount fragments the backward computation into narrow GEMMs that underutilize the
GPU and throttle fine-tuning throughput.
In short, \textit{an iteration-level runtime cannot shorten the 
queueing time without slowing fine-tuning significantly}.

We argue that an inference-prioritized co-serving runtime
should instead decide the backward work during the iteration
rather than before it.
Specifically, it should (a) leave no backward work at the iteration boundary, so that new
requests start as soon as the forward pass ends, while (b) still letting
fine-tuning use the capacity that active requests leave within their SLOs
with microbatches of efficient size. Achieving both requires deciding how much
backward work to run while the inference iteration executes, layer by layer,
which motivates the \name pipeline design as follows.

\section{\name Pipeline}
\label{sec:pipeline}

\subsection{Co-Serving Schedule}
\label{subsec:pipe-frame}
\name divides the backward pass of each fine-tuning microbatch into one
backward task per transformer layer, which applies
Eq.~\ref{eq:lora_grad_x} at that layer. As an inference iteration advances
from layer $1$ to layer $L$, the runtime may execute the inference forward of
layer $l$ with ready backward tasks of layer $l$ from older microbatches.

\definecolor{InfColor}{HTML}{D97925}
\definecolor{FTFwdColor}{HTML}{8CBBD9}
\definecolor{BwdColor}{HTML}{3F6FA5}
\definecolor{BurstColor}{HTML}{B73A2D}
\definecolor{BurstFill}{HTML}{F6E5E2}
\definecolor{MutedColor}{HTML}{626262}

\tikzset{
  work/.style={draw=black, line width=0.45pt, minimum height=4.68mm,
               inner sep=0pt, font=\scriptsize},
  bwd/.style={work, fill=BwdColor, text=white},
  bwdpaused/.style={work, fill=white, draw=MutedColor,
                    densely dashed, text=BwdColor},
  inf/.style={work, fill=InfColor},
  burst/.style={work, fill=BurstFill, text=BurstColor,
                font=\scriptsize\bfseries},
  dep/.style={-{Latex[length=1.4mm,width=1mm]},
              draw=MutedColor, line width=0.55pt},
  timeline/.style={-{Latex[length=1.7mm]},draw=black,line width=0.55pt},
  waitline/.style={draw=MutedColor,densely dotted,line width=0.55pt},
  smalllabel/.style={font=\scriptsize},
  muted/.style={font=\scriptsize},
  done/.style={work,fill=black!10,draw=MutedColor,text=MutedColor},
}

\begin{figure}[t]
\centering
\begin{tikzpicture}[x=1cm,y=1cm]

\draw[MutedColor,densely dashed,line width=.55pt]
  (4.38,3.42) -- (4.38,-.16);

\foreach \x in {1.45,2.28,3.11,3.94,4.82,5.65,6.48,7.31} {
  \draw[waitline] (\x,3.15) -- (\x,.39);
}

\draw[-{Latex[length=1.7mm]},line width=.55pt]
  (1.12,3.15) -- (7.72,3.15);
\node[font=\scriptsize\bfseries,anchor=west]
  at (.02,3.15) {Time};
\foreach \x/\t in {
  1.45/1,2.28/2,3.11/3,3.94/4,
  4.82/5,5.65/6,6.48/7,7.31/8} {
  \draw[line width=.50pt] (\x,3.08) -- (\x,3.22);
  \node[font=\small,anchor=south]
    at (\x,3.24) {$t_{\t}$};
}

\node[font=\scriptsize\bfseries,text=InfColor,anchor=west]
  at (.02,2.70) {FWD};
\foreach \y/\id in {2.27/I,1.84/II,1.41/III,.98/IV,.55/V} {
  \node[muted,anchor=west] at (.02,\y) {BWD-\id};
}

\foreach \x/\l in {1.45/1,2.28/2,3.11/3,3.94/4} {
  \node[inf,minimum width=5.4mm,minimum height=3.3mm]
    (f\l) at (\x,2.70) {$F_{L\l}$};
}
\draw[-{Latex[length=1.2mm]},InfColor,line width=.50pt]
  (f1.east) -- (f2.west);
\draw[-{Latex[length=1.2mm]},InfColor,line width=.50pt]
  (f2.east) -- (f3.west);
\draw[-{Latex[length=1.2mm]},InfColor,line width=.50pt]
  (f3.east) -- (f4.west);

\foreach \x/\l in {4.82/1,5.65/2,6.48/3,7.31/4} {
  \node[inf,minimum width=5.4mm,minimum height=3.3mm]
    (g\l) at (\x,2.70) {$F_{L\l}$};
}
\draw[-{Latex[length=1.2mm]},InfColor,line width=.50pt]
  (g1.east) -- (g2.west);
\draw[-{Latex[length=1.2mm]},InfColor,line width=.50pt]
  (g2.east) -- (g3.west);
\draw[-{Latex[length=1.2mm]},InfColor,line width=.50pt]
  (g3.east) -- (g4.west);

\draw[-{Latex[length=1.2mm]},MutedColor,densely dashed,line width=.42pt]
  (2.28,2.27) -- (1.45,2.27);
\draw[-{Latex[length=1.2mm]},MutedColor,densely dashed,line width=.42pt]
  (3.11,1.84) -- (2.28,1.84);
\draw[-{Latex[length=1.2mm]},MutedColor,densely dashed,line width=.42pt]
  (3.94,1.41) -- (3.11,1.41);

\node[bwd,minimum width=5.2mm,minimum height=3.1mm] (b11) at (1.45,2.27) {$B_{L1}$};
\node[done,minimum width=2.1mm,minimum height=2.1mm] (b12) at (2.28,2.27) {};
\node[done,minimum width=2.1mm,minimum height=2.1mm] (b13) at (3.11,2.27) {};
\node[done,minimum width=2.1mm,minimum height=2.1mm] (b14) at (3.94,2.27) {};

\node[bwdpaused,minimum width=2.1mm,minimum height=2.1mm] (b21) at (1.45,1.84) {};
\node[bwd,minimum width=5.2mm,minimum height=3.1mm] (b22) at (2.28,1.84) {$B_{L2}$};
\node[done,minimum width=2.1mm,minimum height=2.1mm] (b23) at (3.11,1.84) {};
\node[done,minimum width=2.1mm,minimum height=2.1mm] (b24) at (3.94,1.84) {};

\node[bwdpaused,minimum width=2.1mm,minimum height=2.1mm] (b31) at (1.45,1.41) {};
\node[bwdpaused,minimum width=2.1mm,minimum height=2.1mm] (b32) at (2.28,1.41) {};
\node[bwd,minimum width=5.2mm,minimum height=3.1mm] (b33) at (3.11,1.41) {$B_{L3}$};
\node[done,minimum width=2.1mm,minimum height=2.1mm] (b34) at (3.94,1.41) {};

\node[bwdpaused,minimum width=2.1mm,minimum height=2.1mm] (b41) at (1.45,.98) {};
\node[bwdpaused,minimum width=2.1mm,minimum height=2.1mm] (b42) at (2.28,.98) {};
\node[bwdpaused,minimum width=2.1mm,minimum height=2.1mm] (b43) at (3.11,.98) {};
\node[bwd,minimum width=5.2mm,minimum height=3.1mm] (b44) at (3.94,.98) {$B_{L4}$};

\node[bwdpaused,minimum width=2.1mm,minimum height=2.1mm] (b51) at (1.45,.55) {};
\node[bwdpaused,minimum width=2.1mm,minimum height=2.1mm] (b52) at (2.28,.55) {};
\node[bwdpaused,minimum width=2.1mm,minimum height=2.1mm] (b53) at (3.11,.55) {};
\node[bwdpaused,minimum width=2.1mm,minimum height=2.1mm] (b54) at (3.94,.55) {};

\draw[-{Latex[length=1.4mm]},BwdColor,line width=.65pt] (b11) -- (b22);
\draw[-{Latex[length=1.4mm]},BwdColor,line width=.65pt] (b22) -- (b33);
\draw[-{Latex[length=1.4mm]},BwdColor,line width=.65pt] (b33) -- (b44);

\foreach \src/\dst in {
  b14/b13,b13/b12,
  b24/b23,b22/b21,
  b33/b32,b32/b31,
  b44/b43,b43/b42,b42/b41,
  b54/b53,b53/b52,b52/b51} {
  \draw[-{Latex[length=1.2mm]},MutedColor,densely dashed,line width=.42pt]
    (\src.west) -- (\dst.east);
}

\draw[-{Latex[length=1.2mm]},MutedColor,densely dashed,line width=.42pt]
  (5.65,1.84) -- (4.82,1.84);
\draw[-{Latex[length=1.2mm]},MutedColor,densely dashed,line width=.42pt]
  (6.48,1.41) -- (5.65,1.41);
\draw[-{Latex[length=1.2mm]},MutedColor,densely dashed,line width=.42pt]
  (7.31,.98) -- (6.48,.98);

\node[done,minimum width=2.1mm,minimum height=2.1mm] (n11) at (4.82,2.27) {};
\node[done,minimum width=2.1mm,minimum height=2.1mm] (n12) at (5.65,2.27) {};
\node[done,minimum width=2.1mm,minimum height=2.1mm] (n13) at (6.48,2.27) {};
\node[done,minimum width=2.1mm,minimum height=2.1mm] (n14) at (7.31,2.27) {};

\node[bwd,minimum width=5.2mm,minimum height=3.1mm] (n21) at (4.82,1.84) {$B_{L1}$};
\node[done,minimum width=2.1mm,minimum height=2.1mm] (n22) at (5.65,1.84) {};
\node[done,minimum width=2.1mm,minimum height=2.1mm] (n23) at (6.48,1.84) {};
\node[done,minimum width=2.1mm,minimum height=2.1mm] (n24) at (7.31,1.84) {};

\node[bwdpaused,minimum width=2.1mm,minimum height=2.1mm] (n31) at (4.82,1.41) {};
\node[bwd,minimum width=5.2mm,minimum height=3.1mm] (n32) at (5.65,1.41) {$B_{L2}$};
\node[done,minimum width=2.1mm,minimum height=2.1mm] (n33) at (6.48,1.41) {};
\node[done,minimum width=2.1mm,minimum height=2.1mm] (n34) at (7.31,1.41) {};

\node[bwdpaused,minimum width=2.1mm,minimum height=2.1mm] (n41) at (4.82,.98) {};
\node[bwdpaused,minimum width=2.1mm,minimum height=2.1mm] (n42) at (5.65,.98) {};
\node[bwd,minimum width=5.2mm,minimum height=3.1mm] (n43) at (6.48,.98) {$B_{L3}$};
\node[done,minimum width=2.1mm,minimum height=2.1mm] (n44) at (7.31,.98) {};

\node[bwdpaused,minimum width=2.1mm,minimum height=2.1mm] (n51) at (4.82,.55) {};
\node[bwdpaused,minimum width=2.1mm,minimum height=2.1mm] (n52) at (5.65,.55) {};
\node[bwdpaused,minimum width=2.1mm,minimum height=2.1mm] (n53) at (6.48,.55) {};
\node[bwd,minimum width=5.2mm,minimum height=3.1mm] (n54) at (7.31,.55) {$B_{L4}$};

\draw[-{Latex[length=1.4mm]},BwdColor,line width=.65pt] (n21) -- (n32);
\draw[-{Latex[length=1.4mm]},BwdColor,line width=.65pt] (n32) -- (n43);
\draw[-{Latex[length=1.4mm]},BwdColor,line width=.65pt] (n43) -- (n54);

\foreach \src/\dst in {
  n14/n13,n13/n12,n12/n11,
  n24/n23,n23/n22,
  n34/n33,n32/n31,
  n43/n42,n42/n41,
  n54/n53,n53/n52,n52/n51} {
  \draw[-{Latex[length=1.2mm]},MutedColor,densely dashed,line width=.42pt]
    (\src.west) -- (\dst.east);
}

\node[smalllabel] at (2.70,.06) {(a) Iteration~$k$};
\node[smalllabel] at (6.07,.06) {(b) Iteration~$k+1$};

\node[bwd,minimum width=3.8mm,minimum height=2.4mm] at (2.05,-.46) {};
\node[muted,anchor=west] at (2.33,-.46) {ready};
\node[done,minimum width=2.4mm,minimum height=2.4mm] at (3.62,-.46) {};
\node[muted,anchor=west] at (3.83,-.46) {completed};
\node[bwdpaused,minimum width=2.4mm,minimum height=2.4mm] at (5.69,-.46) {};
\node[muted,anchor=west] at (5.90,-.46) {unvisited};

\end{tikzpicture}

\caption{Cross-microbatch backward wavefront.
Each backward microbatch (i.e., BWD-I to BWD-V) follows reverse layer order (gray dashed arrows), while their
next ready tasks (blue) collectively follow the forward traversal.}
\label{fig:backward-wavefront}
\end{figure}
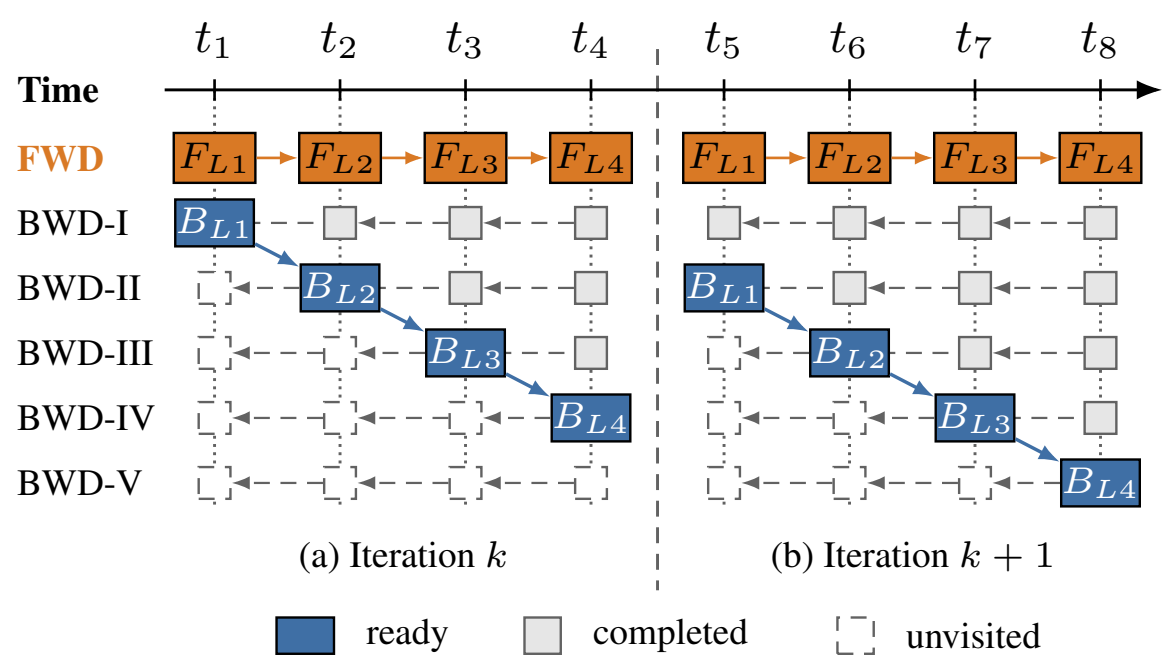

\noindent\textbf{Key properties.}
When the forward pass finishes the last layer $L$, this iteration ends 
without any outstanding backward task, so a request that arrived during the iteration is
admitted immediately, which makes our schedule \emph{inference first}.
The runtime adds a backward task at a layer only if the task fits within the
latency headroom of the active inference requests.
During a burst, the runtime withholds
backward tasks from every layer that it has not yet dispatched, which makes the backward tasks \emph{opportunistic}.

Fig.~\ref{fig:pipeline-modes} compares our schedule with the three FlexLLM
modes for a request that arrives during an iteration. In every FlexLLM mode,
the request also waits for the backward work attached to the iteration.
Whereas in \name, it waits only for the rest of the inference forward pass and
the backward tasks already dispatched, since the runtime withholds backward
tasks from the remaining layers.

\noindent\textbf{Backward Wavefront.}
A single microbatch cannot feed this schedule, because its backward tasks
become ready in reverse layer order while inference visits the layers in
forward order. \name therefore keeps a \emph{wavefront} of microbatches whose
next tasks are staggered by one layer, so that every layer of an inference
iteration finds a ready task (Fig.~\ref{fig:backward-wavefront}). Each
iteration advances every microbatch by one layer, after which the oldest one
completes and a new one enters at the last layer, restoring the same pattern.
The runtime tracks the wavefront with a FIFO queue per layer and the 
backward pass of each microbatch spreads over at least $L$ successive iterations.

\subsection{SLO-Aware Scheduling}
\label{sec:design_scheduling}

We propose an SLO-aware scheduler in \name that decides how many ready
backward tasks to run before each layer. The scheduler follows a simple greedy
rule based on the latency headroom of the active inference requests.

The latency headroom $H_l$ is the time that backward tasks may still add to
the current iteration at layer $l$ without making any request miss its SLO.
To calculate $H_l$, the scheduler (1) takes the time remaining until the
earliest TTFT or TPOT deadline among unfinished requests, (2) subtracts the
inference-only latency of the layers that remain in the iteration, i.e.,
layers $l$ to $L$, and (3) if some requests are still waiting for their first
token, subtracts the inference-only time to prefill them as well.

To estimate how much of $H_l$ the backward tasks consume, the scheduler uses
a profiled latency model $C(n_I,k)$. Given the number of inference tokens
$n_I$ in the current batch and a number of backward tasks $k$, the model
returns the latency that the $k$ backward tasks add to the current layer,
measured as the median increase over the same layer without backward tasks.

With these two quantities, the greedy rule picks the largest $k_l$ such that
$C(n_I,k_l)\le H_l$ and runs the backward tasks of the $k_l$ oldest
microbatches queued at layer $l$, since they are closest to completing their
backward pass. We prefer this greedy rule to planning all remaining layers jointly, because
a backward task costs roughly the same at every layer of an iteration. Saving
headroom for later layers would therefore not let more tasks run, whereas
spending it immediately finishes the oldest microbatches first.

Finally, \name accumulates the gradients of multiple microbatches that all
run against the same LoRA adapter weights, and it applies the optimizer update
only after the last of them finishes its backward pass. Before the next group
of microbatches starts, this optimizer barrier drains the wavefront, which
takes at least $L$ iterations, since a microbatch advances by one layer per
iteration.

\section{Bounding Memory and Fusing Shared-Weight Computation}
\label{sec:memory-kernel}

Realizing the above pipeline exposes two new challenges. First, the wavefront
keeps many microbatches paused in the middle of their backward passes, and
each of them retains activations for the layers it has yet to process.
Second, the pipeline pairs the inference forward and the fine-tuning backward
on the same base weight, whose two GEMMs reduce along opposite dimensions. We address the first challenge with scheduler-guided offloading in
\S\ref{subsec:design-offload} and the second with a shared-weight GEMM
kernel in \S\ref{subsec:design-kern}.

\subsection{Scheduler-Guided Offloading}
\label{subsec:design-offload}

Suppose that each layer retains $a$ bytes of activations per microbatch and
that each microbatch holds a boundary gradient of $g$ bytes. Iteration-level
co-serving, e.g., FlexLLM, runs the backward pass of one microbatch at a time and thus
retains at most $La+g$ bytes. A full wavefront, in contrast, holds one
microbatch at each layer $r$, which still retains the activations of layers
$1,\ldots,r$, so it occupies
\begin{equation}
    M_{\mathrm{wave}} =
    \sum_{r=1}^{L}(ra+g) = \frac{L(L+1)}{2}a+Lg.
    \label{eq:full_wavefront_memory}
\end{equation}
For FP16 Llama-3.1-8B, where $a$ is about 120~KiB per token, $M_{\mathrm{wave}}$ reaches 124.4~GiB
for 2048-token microbatches, compared with 7.5~GiB under iteration-level
co-serving. Keeping this state in GPU memory could be impractical.

\name therefore offloads the retained state to pinned host memory. As soon as
the fine-tuning forward pass completes a layer, the offload manager copies the
retained activations of that layer to a preallocated host memory region on a
separate CUDA stream. It also offloads each boundary gradient right after the
backward task that produces it. As a result, the GPU holds only the state of
the microbatches that are currently being produced or consumed.
When the host PCIe link becomes the offloading bottleneck, 
e.g., while training for long sequences,
recomputing cheap tensors instead of offloading them could further reduce
this traffic.

\name additionally ensures that offloading does not stall computation by
prefetching each microbatch's state back to the GPU before its backward task
runs. Since the scheduler's plan already determines which microbatches each
upcoming layer will run, the offload manager uses this plan to time the
prefetches. For each planned microbatch, it walks back
from the layer at which the microbatch will run, accumulating profiled layer
times until they cover the host-to-device transfer, and issues the prefetch
at that layer, serving the earliest use first. When a new incoming request revises 
the scheduler's plan, both the old and new plans select FIFO
prefixes of each layer's queue, so the offload manager keeps the prefetches they share
and fetches only the newly selected state.

\subsection{Shared-Weight Forward-Backward Kernel}
\label{subsec:design-kern}

\begin{figure}[t]
\centering
\captionsetup[subfigure]{
labelfont=bf,
  justification=centering,singlelinecheck=true,skip=8pt}

\begin{subfigure}[t]{0.38\columnwidth}
\centering
\resizebox{\linewidth}{!}{%
\begin{tikzpicture}[
  x=1cm,y=1cm,
  tile/.style={draw=black,line width=.45pt,minimum width=6.2mm,
               minimum height=5.0mm,inner sep=0pt,font=\scriptsize},
  fwd/.style={-{Latex[length=1.35mm,width=.9mm]},draw=InfColor,
              line width=.72pt},
  boundary/.style={draw=MutedColor!72,
                   dash pattern=on 2.2pt off 1.6pt,line width=.42pt},
  muted/.style={font=\scriptsize,text=MutedColor}]

\node[font=\small] at (1.88,3.27)
  {$Y=\textcolor{InfColor}{W}X$};

\foreach \r/\y/\rid in {1/2.60/1,2/2.00/2,3/1.40/3} {
  \node[muted,anchor=east] at (.62,\y) {CTA $\rid$};
  \foreach \c/\x/\cid in {1/1.00/1,2/1.68/2,3/2.36/3} {
    \ifnum\r=2
      \node[tile,fill=InfColor!16] (a\r\c) at (\x,\y)
        {$W_{\rid\cid}$};
    \else
      \node[tile,fill=black!3] (a\r\c) at (\x,\y)
        {$W_{\rid\cid}$};
    \fi
  }
}
\draw[boundary] (.69,2.30) -- (3.58,2.30);
\draw[boundary] (.69,1.70) -- (3.58,1.70);
\node[draw=InfColor,fill=white,text=InfColor,line width=.55pt,
      minimum width=7mm,minimum height=4.2mm,inner sep=0pt,
      font=\scriptsize] (yi) at (3.23,2.00) {$Y_2$};
\draw[fwd] (a23.east) -- (yi.west);

\end{tikzpicture}%
}
\caption{Forward reduction.}
\label{fig:orthogonal-accumulation-fwd}
\end{subfigure}%
\hfill
\begin{subfigure}[t]{0.60\columnwidth}
\centering
\resizebox{\linewidth}{!}{%
\begin{tikzpicture}[
  x=1cm,y=1cm,
  tile/.style={draw=black,line width=.45pt,minimum width=6.2mm,
               minimum height=5.0mm,inner sep=0pt,font=\scriptsize},
  bwd/.style={-{Latex[length=1.35mm,width=.9mm]},draw=BwdColor,
              line width=.72pt},
  boundary/.style={draw=MutedColor!72,
                   dash pattern=on 2.2pt off 1.6pt,line width=.42pt},
  muted/.style={font=\scriptsize,text=MutedColor}]

\node[font=\small] at (2.80,3.27)
  {$\nabla_X\mathcal{L}=\textcolor{BwdColor}{W^{\mathsf T}}G$};

\foreach \r/\y/\rid in {1/2.60/1,2/2.00/2,3/1.40/3} {
  \node[muted,anchor=east] at (.80,\y) {CTA $\rid$};
  \foreach \c/\x/\cid in {1/1.18/1,2/1.86/2,3/2.54/3} {
    \ifnum\c=2
      \node[tile,fill=BwdColor!16] (b\r\c) at (\x,\y)
        {$W_{\rid\cid}$};
    \else
      \node[tile,fill=black!3] (b\r\c) at (\x,\y)
        {$W_{\rid\cid}$};
    \fi
  }
  \node[draw=BwdColor,fill=white,text=BwdColor,line width=.55pt,
        minimum width=8mm,minimum height=4.2mm,inner sep=0pt,
        font=\scriptsize] (p\r) at (3.45,\y) {$P_{\rid 2}$};
  \draw[bwd] (b\r3.east) -- (p\r.west);
  \draw[BwdColor,line width=.72pt] (p\r.east) -- (3.96,\y);
}
\draw[boundary] (.87,2.30) -- (3.80,2.30);
\draw[boundary] (.87,1.70) -- (3.80,1.70);
\draw[BwdColor,line width=.72pt] (3.96,2.60) -- (3.96,1.40);
\node[draw=BwdColor,fill=BwdColor!12,text=BwdColor,line width=.60pt,
      circle,minimum size=5.4mm,inner sep=0pt,
      font=\scriptsize] (reduce) at (4.42,2.00) {$\Sigma$};
\draw[bwd] (3.96,2.00) -- (reduce.west);
\node[draw=BwdColor,fill=white,text=BwdColor,line width=.55pt,
      minimum width=9mm,minimum height=4.2mm,inner sep=0pt,
      font=\scriptsize] (dxj) at (5.28,2.00) {$\nabla_{X_2}\mathcal{L}$};
\draw[bwd] (reduce.east) -- (dxj.west);

\end{tikzpicture}%
}
\caption{Cross-CTA backward reduction.}
\label{fig:orthogonal-accumulation-bwd}
\end{subfigure}

\caption{A row-wise weight partition completes forward locally but distributes backward across CTAs.}
\label{fig:orthogonal-accumulation}
\end{figure}
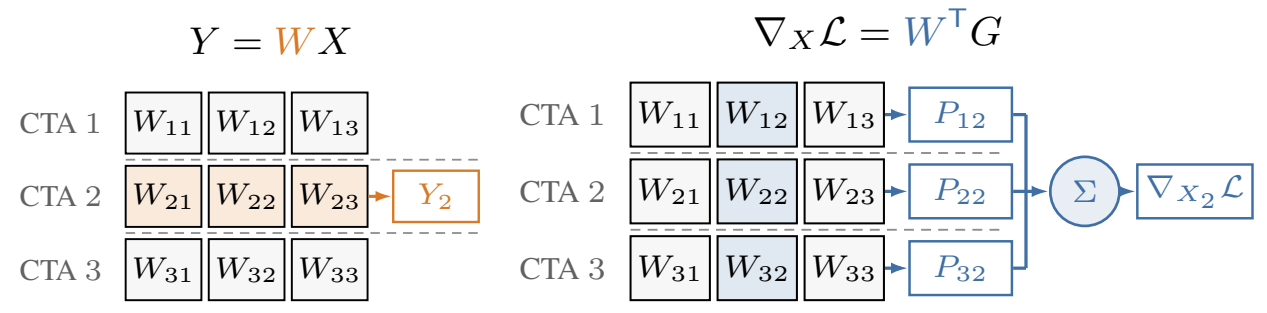

As established in \S\ref{sec:background_lora}, the inference forward computes $WX$ while the
fine-tuning backward computes $W^{\mathsf T}G$.
A fused kernel can load each tile of $W$ once
for both GEMMs and thereby halve the base weight reads.

However, realizing this saving is not straightforward.
A GEMM kernel assigns output tiles to thread blocks
(CTAs)~\cite{nvidia_cutlass_efficient_gemm}, and since the two GEMMs reduce
along opposite dimensions of $W$, a weight partition that completes one output
within a CTA leaves partial sums of the other output spread across CTAs, 
as illustrated in Fig.~\ref{fig:orthogonal-accumulation}. If the wide backward output is the
one spread across CTAs, the reduction traffic erases the saving. For instance, for
\texttt{down\_proj} with 64-row panels and 128 backward tokens, a
row-wise mapping writes about 470~MB of FP32 partial gradients to save a
117~MB weight read, which outweighs the saving four times over.

Since the two GEMMs reduce along different dimensions, one of their outputs
must be reduced across CTAs. On commodity GPUs, this reduction can be costly, because
CTAs cannot share registers or shared memory, so their partial sums must
travel through global memory\footnote{NVIDIA Hopper's distributed shared memory lets the CTAs of a
thread-block cluster reduce partial sums in each other's shared memory, but
the cost of cross-CTA reductions remains substantial compared with
reductions in local shared memory.}. \name therefore assigns this cross-CTA reduction to the inference output,
which spans far fewer tokens than the backward output. It partitions $W$ into
column panels, and each CTA processes all rows of its panel, so the backward
output of a panel accumulates within a single CTA, while the CTAs add their
partial inference outputs to a shared FP32 buffer with atomic operations.

We further improve Tensor Core utilization in the fused kernel with a
software pipeline shared by both GEMMs. Specifically, we design a
three-stage \texttt{cp.async} pipeline that prefetches the next tiles
of $W$ and $G$ into shared memory while the current tile is being computed,
which hides HBM latency, and maintains a swizzled layout that avoids shared-memory bank
conflicts. Each staged weight tile is consumed by both GEMMs before its
stage is refilled. The fused kernel reads the tile into registers with
\texttt{ldmatrix}, using its transposing variant for the backward GEMM, so
both GEMMs obtain the operand layouts of the Tensor Cores without extra
copies, and both execute \texttt{mma.sync.m16n8k16} Tensor Core instructions.
The inference input $X$ stays in registers for a whole panel.
Finally, the kernel launches one persistent CTA per SM and
assigns each CTA the same number of weight rows to maximize SM utilization, following~\cite{osama2023streamk}.

\section{Implementation}
\label{sec:impl}

We implement \name on top of FlexLLM~\cite{oliaro2026flexllm}, which builds on
FlexFlow Serve and the Legion runtime~\cite{bauer2012legion}, with 2.9K
lines of C++ and CUDA. We leave the operators, LoRA kernels, and optimizer of
FlexLLM unchanged for fair comparison.

\begin{figure}[t]
\centering
\includegraphics[width=.9\columnwidth,trim=24pt 14pt 24pt 13pt,clip]{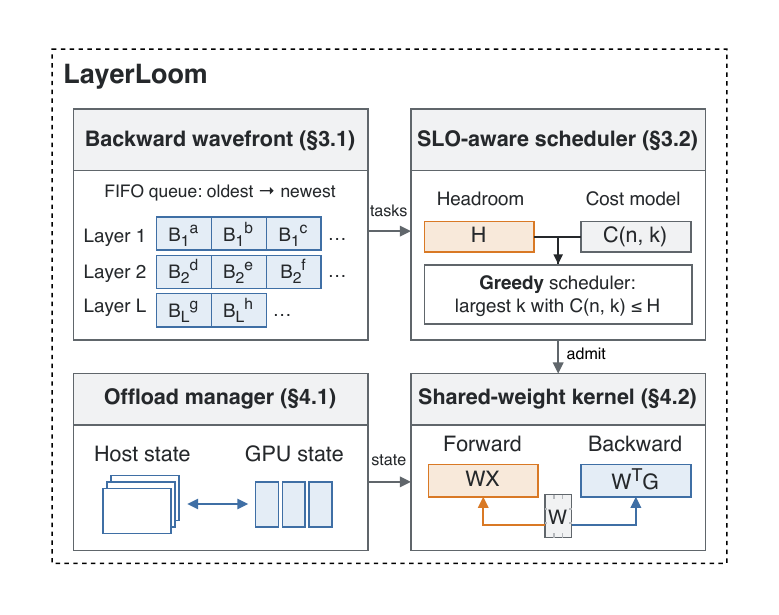}
\caption{\name system architecture.}
\label{fig:system-overview}
\vspace{-2mm}
\end{figure}

Fig.~\ref{fig:system-overview} shows the architecture of the \name system.
The backward wavefront (\S\ref{subsec:pipe-frame}) organizes ready tasks in per-layer FIFO queues, and
the SLO-aware scheduler (\S\ref{sec:design_scheduling}) selects how many to run using latency headroom and
a profiled cost model. The offload manager (\S\ref{subsec:design-offload}) keeps retained state in host
memory and prefetches it for the selected tasks. The shared-weight kernel (\S\ref{subsec:design-kern})
fuses the inference-forward and fine-tuning-backward GEMMs to reuse each
base-weight tile.

\raggedbottom
\section{Evaluation}
\label{sec:eval}

\subsection{Experimental Setup}
\label{subsec:eval-setup}

Our evaluation runs on one NVIDIA RTX 4090 GPU with Llama-3.1-8B in FP16,
and we apply rank-16 LoRA to \texttt{down\_proj}. Inference requests replay
the high-burstiness trace of \S\ref{subsec:motivation} at two loads, which we calibrate by the joint SLO attainment of
inference alone so that neither load overloads the GPU by itself.
Specifically, inference alone attains 100\% at the \emph{Mid} load and 98\%
at the \emph{High} load.

In every evaluated system, fine-tuning uses microbatches of 128 tokens and accumulates
gradients over 64 microbatches per optimizer update. A request meets its SLO
if its TTFT is at most 5~s and its TPOT is at most 168~ms, which is
2.5$\times$ the median TPOT of inference alone on this GPU. We report (1) the
joint SLO attainment, i.e., the fraction of requests that meet both TTFT and TPOT targets,
and (2) the fine-tuning (FT) throughput in trained tokens per second.

We compare \name with the following baselines.
\begin{itemize}
  \item \textbf{Inference-only} runs FlexLLM without fine-tuning and gives
  the SLO attainment that other systems should approach.
  \item \textbf{FlexLLM} (FLLM) runs its co-serving mode with the default
  backward setting.
  \item \textbf{FlexLLM-temporal} (FLLM-T) and \textbf{FlexLLM-spatial}
  (FLLM-S) run FlexLLM's temporal and spatial sharing modes with their
  default settings.
  \item \textbf{FlexLLM-Backward-$b$} (FLLM-B$b$) limits co-serving to $b$
  backward layers per iteration, which represents the strategy of attaching
  less backward work to each iteration, and $b=1$ is its most responsive
  setting.
\end{itemize}
All systems run the same cuBLAS kernels without Legion
tracing~\cite{bauer2012legion}, because the per-layer graphs of \name change
with every scheduling decision and cannot be replayed as one fixed trace.
Unless noted otherwise, \name uses the greedy scheduler of
\S\ref{sec:design_scheduling} and a wavefront of at most 16
microbatches, and it enables offloading only when GPU memory is insufficient.
\name and all baselines use the same configuration at both loads.
Each experiment is
repeated at least three times, and we report the mean results.

\subsection{End-to-End SLO and Throughput}
\label{subsec:eval-main}

\begin{figure}[t]
\centering
\includegraphics[width=\columnwidth]{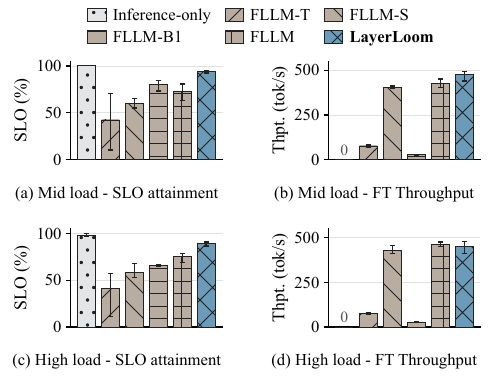}
\caption{Joint SLO attainment and fine-tuning throughput at the Mid (a, b) and
High (c, d) loads.}
\label{fig:eval-main}
\vspace{-2mm}
\end{figure}

We first compare the end-to-end SLO attainment and fine-tuning throughput of
\name with those of the baselines, and we make two observations.

First, \name achieves higher SLO attainment than FlexLLM at both loads
without sacrificing fine-tuning throughput, as shown in
Fig.~\ref{fig:eval-main}. At the Mid load, \name attains 94.4\% joint SLO
attainment compared with 72.6\% for FlexLLM, and it also trains 12\% more
tokens per second, i.e., 475 compared with 425. The High load shows a similar
advantage, where \name attains 88.8\% compared with 75.2\% for FlexLLM while
training a comparable 449 versus 463 tokens per second. In summary, \name
improves SLO attainment by 21.8 and 13.5 points at the two loads while
keeping fine-tuning throughput on par with or above FlexLLM.

Second, \name outperforms all other FlexLLM variants by at least 14.1 points in SLO attainment at both loads, while none of them trains more tokens per
second. Specifically, (1) FLLM-B1, which attaches only one backward layer to each iteration,
attains only 80.3\% and 66.0\% SLO attainment while training 28 and 27 tokens
per second at the two loads, i.e., 16--17$\times$ less than \name. (2) FLLM-T performs worst, with
42.3\% and 41.3\% SLO attainment at about 77 tokens per second, because every
request that arrives during a standalone fine-tuning pass waits for the whole
pass. (3) FLLM-S trains 404 and 428 tokens per second but attains only 60.3\% and
58.3\% at the two loads, since the backward stream slows the forward pass and the next batch
still waits for both streams. \name, in contrast, batches the backward tasks
of each layer and withholds them only when the headroom is short, which keeps
both its SLO attainment and its throughput high.

\begin{figure}[t]
\centering
\captionsetup[subfigure]{font=footnotesize,skip=1pt}
\includegraphics[width=\columnwidth]{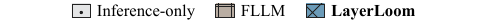}\par\vspace{-3pt}
\begin{subfigure}[t]{0.49\columnwidth}
\centering
\includegraphics[width=\linewidth]{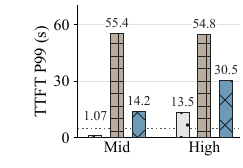}
\caption{TTFT}
\label{fig:eval-latency-ttft}
\end{subfigure}\hfill
\begin{subfigure}[t]{0.49\columnwidth}
\centering
\includegraphics[width=\linewidth]{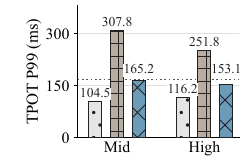}
\caption{TPOT}
\label{fig:eval-latency-tpot}
\end{subfigure}
\caption{Inference latency. Dotted lines mark the SLO targets.}
\label{fig:eval-latency}
\vspace{-2mm}
\end{figure}

\noindent\textbf{Tail latency.}
To understand where the higher SLO attainment of \name comes from, we further
compare the P99 TTFT and TPOT of inference alone, FlexLLM, and \name in
Fig.~\ref{fig:eval-latency}, and we make two observations.

First, \name lowers the P99 TTFT from 55.4~s to 14.2~s at Mid and from
54.8~s to 30.5~s at High, because a request that arrives during a burst no
longer waits for backward work that was committed when the iteration started.
It also keeps the P99 TPOT at 165~ms and 153~ms, below the 168-ms target,
whereas FlexLLM reaches 308~ms and 252~ms, respectively.

Second, \name still trails inference alone by 5.6 points at Mid and 9.2 points
at High. Since its P99 TPOT meets the target, the remaining violations come
from the first token. Its P99 TTFTs of 14.2~s and 30.5~s show that requests still queue
during the largest bursts, when the 16-request batch is full, and the observed
iteration latency exceeds the scheduler's prediction in 18--23\% of
iterations, so the latency model occasionally admits more backward tasks than
the headroom allows.

\subsection{Wavefront Size and Memory Offloading}
\label{subsec:eval-memory}

Next, we evaluate how the offload manager of \name enables larger wavefronts
and saves GPU memory, and how the wavefront size affects the two metrics.

\begin{figure}[t]
\centering
\includegraphics[width=\columnwidth]{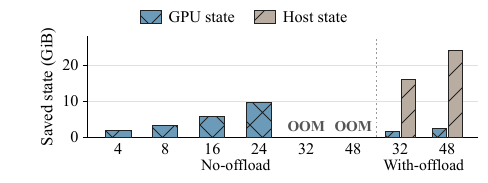}
\caption{Saved fine-tuning state at the Mid load in GPU and pinned host
memory. OOM marks configurations that run out of GPU memory.}
\label{fig:eval-memory}
\vspace{-2mm}
\end{figure}

Fig.~\ref{fig:eval-memory} shows how the saved fine-tuning state grows with
the wavefront at the Mid load.
Without offloading, \name can only scale to 24 microbatches, and the
runs with 32 and 48 microbatches run out of GPU memory. The saved state grows
from 1.8~GiB with 4 microbatches to 9.7~GiB with 24, while the GPU also holds the LLM weights and the KV
cache.
With the offload manager, \name places the saved state of 32 and 48
microbatches in 16.0~GiB and 24.0~GiB of pinned host memory, respectively.
Only 1.6--2.4~GiB of state remains on the GPU for staging and
prefetching.

\begin{figure}[t]
\centering
\includegraphics[width=\columnwidth]{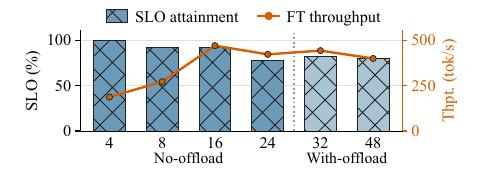}
\caption{Joint SLO attainment (bars) and fine-tuning throughput (line) versus
the wavefront limit $M$ at the Mid load, from single 300-s runs.}
\label{fig:eval-tradeoff}
\vspace{-2mm}
\end{figure}

To examine how the wavefront size affects fine-tuning throughput and
SLO attainment, we vary the wavefront limit from 4 to 48 microbatches at the
Mid load, enabling offloading for 32 and 48 microbatches, and show the results in Fig.~\ref{fig:eval-tradeoff}.

Fine-tuning throughput rises from 187 to 470 tokens per second as the
wavefront grows from 4 to 16 microbatches, while SLO attainment falls only
from 100\% to 92\%, because a larger wavefront places ready backward tasks at
more layers and lets the scheduler spend more of the headroom. From 24 microbatches on, both metrics level off at about 80\% SLO attainment
and 400--440 tokens per second, including the runs with 32 and 48
microbatches that require offloading, which suggests that the headroom rather
than the number of ready tasks then limits the backward work.

In summary, offloading allows the wavefront to grow beyond GPU memory, and the wavefront size provides a knob that trades SLO attainment for
fine-tuning throughput.

\subsection{Scheduling Policy}
\label{subsec:eval-policy}

In \S\ref{sec:design_scheduling}, we argued that the greedy rule of \name
needs neither a plan for the remaining layers nor headroom saved for them. To
test both arguments, we compare the greedy rule with two alternative policies
that we also implement in \name. We run all three policies with 48
fine-tuning microbatches and offloading enabled, since this setting has the
most microbatches competing for the headroom.

(1) The \emph{lookahead} policy plans all remaining layers of the iteration at
once. It chooses the number of tasks at each layer so that their total cost
fits within the headroom, favoring layers whose queue is longer than that of
the next layer in the spirit of MaxWeight
scheduling~\cite{tassiulas1992stability}, and it solves this choice by dynamic
programming before every layer.
(2) The \emph{split} policy instead saves headroom for later layers by giving
each remaining layer a share of the headroom proportional to its queue length.

\begin{table}[t]
\centering
\caption{Scheduling policies of \name with 48 backward
\mbox{microbatches}, averaged over five runs.}
\label{tab:eval-policy}
\small
\setlength{\tabcolsep}{5pt}
\begin{tabular}{@{}lrrrr@{}}
\toprule
& \multicolumn{2}{c}{Mid load} & \multicolumn{2}{c}{High load} \\
\cmidrule(lr){2-3}\cmidrule(lr){4-5}
Policy & SLO (\%) & FT (tok/s) & SLO (\%) & FT (tok/s) \\
\midrule
\bf Greedy (ours) & \bf 88.7 & \bf 511 & \bf 81.9 & \bf 478 \\
Lookahead & 86.9 & 463 & 77.3 & 437 \\
Split     & 78.7 & 485 & 79.8 & 466 \\
\bottomrule
\end{tabular}
\end{table}

First, Table~\ref{tab:eval-policy} shows that the greedy rule attains the
highest SLO attainment and fine-tuning throughput at both loads.
Specifically, it attains 1.8 and 4.6 points higher SLO attainment and 10\%
and 9\% higher throughput than the lookahead policy at the Mid and High
loads, and 10.0 and 2.1 points higher SLO attainment and 5\% and 3\% higher
throughput than the split policy.
Second, to understand why the greedy rule performs better, we replay 8,192
recorded scheduling decisions. Among the decisions in which a queued task fits
within the headroom of the current layer, the greedy rule schedules a backward
task every time, whereas the lookahead policy skips 14\% of them because its
plan reserves headroom for later layers, and the split policy skips 35\%
because its share of the headroom often falls below the cost of a single task.
These results suggest that holding headroom back for later layers, whether by
planning or by splitting, leaves headroom unused without letting more tasks
run.
Lastly, the greedy rule also has the lowest runtime overhead, taking
0.1~$\mu$s per decision, compared with 48~$\mu$s for the lookahead policy and
0.14~$\mu$s for the split policy.

\subsection{Shared-Weight Kernel}
\label{subsec:eval-kernel}

\begin{figure}[t]
\centering
\includegraphics[width=\columnwidth]{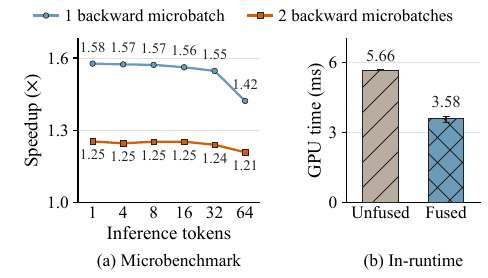}
\caption{Shared-weight \texttt{down\_proj} kernel. (a) Speedup over two
unfused kernels on the same mainloop. (b) \texttt{down\_proj} GPU time per
iteration inside \name at the Mid load.}
\label{fig:eval-kernel}
\vspace{-2mm}
\end{figure}

Next, we evaluate the shared-weight kernel by comparing it with two unfused
kernels, one for $WX$ and one for $W^{\mathsf T}G$, which isolates the effect 
of reading each base weight tile only
once.

We run the fused kernel
and the unfused pair on the \texttt{down\_proj} shape of
Llama-3.1-8B, with 1--64 inference tokens and one or two backward
microbatches, i.e., 128 or 256 backward tokens.
As shown in Fig.~\ref{fig:eval-kernel}(a), the fused kernel is 1.42--1.58$\times$ faster
than the unfused pair for one backward microbatch and 1.21--1.25$\times$
faster for two. The speedup decreases with more backward microbatches, because the saved weight read
becomes a smaller share of the work.

To further measure the benefit inside the full runtime, we run \name with
either kernel at the Mid load, with a wavefront of 24 microbatches, and measure the time of every \texttt{down\_proj} invocation with CUDA events.
The fused kernel reduces the GPU time of \texttt{down\_proj} from 5.66~ms to 3.58~ms per
iteration, a 1.58$\times$ reduction, as shown in
Fig.~\ref{fig:eval-kernel}(b).

In summary, reading each base weight once for both GEMMs speeds up
\texttt{down\_proj} by up to 1.58$\times$, both in isolation and inside the
full runtime.

\subsection{Training Correctness}
\label{subsec:eval-correct}

\begin{figure}[t]
\centering
\includegraphics[width=\columnwidth]{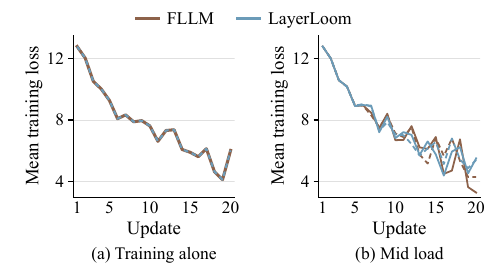}
\caption{Mean training loss per optimizer update (a) without inference load,
where the two curves overlap, and (b) under the Mid load, with two runs per
system.}
\label{fig:eval-loss}
\vspace{-2mm}
\end{figure}

To verify the training correctness of \name, we feed \name and FlexLLM the same samples in the same order with the same learning rate, and we log the
loss of every sample together with its adapter version, i.e., the number of
updates applied at each of the 32 layers when its forward pass ran.

Without inference load, \name and FlexLLM produce the same losses for all
1,280 samples of 20 updates, with a maximum relative difference of
$3.1\times10^{-5}\%$, as shown in Fig.~\ref{fig:eval-loss}(a). The wavefront,
the state transfers, and the optimizer barrier therefore preserve sequential
gradient accumulation.

Co-serving makes the training numerics of both systems depend on which
inference tokens share each GEMM with the fine-tuning tokens. Before the first update, when all forward passes use identical adapter
weights, the per-sample losses of both \name and FlexLLM under the Mid load
match training alone within 0.034\%, so a co-served forward pass computes the
same loss up to FP16 rounding.
After the first update, these small rounding differences change
the gradients slightly and accumulate, so even two runs of FlexLLM under the
same load diverge, with a median per-sample loss difference of 0.15. The loss
of \name differs from FlexLLM's by a median of 0.12--0.38 across pairs of
runs, which is of the same order, and the per-update curves of all four runs
follow the same trend in Fig.~\ref{fig:eval-loss}(b).

In summary, \name trains exactly like FlexLLM without inference load, and
under load its losses deviate from FlexLLM's by the same order as two runs of
FlexLLM deviate from each other.
\flushbottom

\section{Discussion and Future Work}
\label{sec:discussion}

\noindent\textbf{Trace-compatible execution.}
Our prototype runs all systems without Legion tracing, because the per-layer
task graphs of \name change with every scheduling decision. 
However, since the set of
distinct layer graphs is small, i.e., one graph per <inference size, number
of backward tasks> combination, each graph could be captured once as a
Legion trace and replayed thereafter. We take such
trace-compatible execution to raise the throughput of \name as future work.

\noindent\textbf{More GPUs, models, and kernels.}
Extending the evaluation to other GPUs, larger models,
multi-GPU serving, and LoRA on more projections is also future work. In
particular, our shared-weight kernel could be improved by porting it
to NVIDIA Hopper and newer GPUs, whose Tensor Memory Accelerator (TMA) and
warpgroup-level MMA instructions change how weight tiles are staged and
multiplied.
We expect the benefit of \name to hold in these settings.
This is because (1) the problem \name
solves, i.e., inference waiting for backward work committed at the start of
each iteration, arises from how existing runtimes schedule fine-tuning, not
from a particular GPU or model, and (2) \name profiles its latency model on each
target GPU and model, so its scheduler adapts to their speed.

\noindent\textbf{Comparison with GPU preemption.}
GPU preemption systems~\cite{ng2023paella,chen2026deltaserve}
reduce the wait of inference by stopping already issued work, whereas \name
largely avoids issuing backward work that would delay inference in the first
place.
These preemptive approaches are complementary to \name, and a comparison with them is future work.

\section{Related Work}
\label{sec:related}

\noindent\textbf{Co-serving inference and fine-tuning.}
FlexLLM~\cite{oliaro2026flexllm} merges fine-tuning tokens
into the inference forward pass and attaches backward work to each serving
iteration. Other systems multiplex tuning and serving across
GPUs~\cite{he2025resource}, unify multi-LoRA fine-tuning and serving in one
virtualized model~\cite{zhang2026loquetier}, harvest idle cycles through
fine-grained preemption~\cite{qiao2024conserve}, or suspend backward work at
layer boundaries when inference demand appears~\cite{chen2026deltaserve}.
These systems decide when backward work may run or must stop. \name addresses
a complementary question, namely which backward tasks are ready at each layer
of an inference iteration, so that inference finds useful fine-tuning work
during its forward pass.

\noindent\textbf{GPU sharing and LLM serving.}
GPU-sharing systems colocate independent
jobs~\cite{yu2020fine,strati2024orion,zhang2025improving,wang2025colocating}
and preempt them at fine
granularity~\cite{bai2020pipeswitch,han2022microsecond,ng2023paella}, but they
treat the colocated jobs as independent graphs and cannot share their weights
or computation. LLM serving systems reduce inference latency through
continuous batching~\cite{yu2022orca}, paged KV memory~\cite{kwon2023efficient},
chunked prefill~\cite{agrawal2024taming}, preemptive
scheduling~\cite{wu2026fastserve}, disaggregation and
migration~\cite{zhong2024distserve,sun2024llumnix}, layer-level
scheduling~\cite{lee2026tokens}, and multi-LoRA
batching~\cite{chen2024punica,sheng2024slora,wu2024dlora,xue2026muxtune}.
These techniques are orthogonal to \name, which inherits the batching of
FlexLLM.

\noindent\textbf{Pipelined training and memory reduction.}
Pipeline-parallel training interleaves microbatches across the stages of
several GPUs~\cite{huang2019gpipe,narayanan2019pipedream}. The wavefront of
\name similarly interleaves microbatches, but across the layers of one GPU and
clocked by the inference forward pass, and its gradient accumulation with an
optimizer barrier gives it the synchronous semantics of GPipe. Activation
recomputation~\cite{chen2016training} and
offloading~\cite{rhu2016vdnn,ren2021zerooffload} trade computation or host
bandwidth for GPU memory, and \name applies offloading to the wavefront while
timing its prefetches with the plan of the scheduler.

\section{Conclusion}
\label{sec:conclusion}

Co-serving LLM inference with LoRA fine-tuning can turn the capacity that
bursty inference leaves idle into fine-tuning progress, but existing runtimes
fix the backward work before each serving iteration and make new requests
wait for it. This paper presented \name, which instead anchors backward 
tasks to the inference forward pass at layer granularity. A wavefront of fine-tuning
microbatches supplies ready backward tasks in forward layer order, an
SLO-aware scheduler decides at every layer how many of them to run,
offloading keeps the state of the wavefront off the GPU, and a shared-weight
kernel reads each base weight once for both workloads. On an RTX 4090 GPU
with the Llama-3.1-8B architecture, \name achieves 13.5--21.8 points higher
joint SLO attainment than FlexLLM with comparable or up to 12\% higher fine-tuning
throughput, outperforms FlexLLM's temporal and spatial sharing modes, and
matches the training losses of FlexLLM without inference load.



\bibliography{ref}

@inproceedings{hu2022lora,
  title = {{LoRA: Low-Rank Adaptation of Large Language Models}},
  author = {Hu, Edward J. and Shen, Yelong and Wallis, Phillip and Allen-Zhu, Zeyuan and Li, Yuanzhi and Wang, Shean and Wang, Lu and Chen, Weizhu},
  booktitle = {International Conference on Learning Representations},
  year = {2022}
}

@inproceedings{wang2025burstgpt,
  title = {{BurstGPT: A Real-World Workload Dataset to Optimize LLM Serving Systems}},
  author = {Wang, Yuxin and Chen, Yuhan and Li, Zeyu and Kang, Xueze and Fang, Yuchu and Zhou, Yeju and Zheng, Yang and Tang, Zhenheng and He, Xin and Guo, Rui and Wang, Xin and Wang, Qiang and Zhou, Amelie Chi and Chu, Xiaowen},
  booktitle = {Proceedings of the 31st ACM SIGKDD Conference on Knowledge Discovery and Data Mining V.2},
  pages = {5831--5841},
  year = {2025}
}

@inproceedings{he2025resource,
  title = {{Resource Multiplexing in Tuning and Serving Large Language Models}},
  author = {Yongjun He and Haofeng Yang and Yao Lu and Ana Klimovic and Gustavo Alonso},
  booktitle = {2025 USENIX Annual Technical Conference},
  pages = {1639--1655},
  year = {2025}
}

@inproceedings{zhang2026loquetier,
  title = {{Loquetier: A Virtualized Multi-LoRA Framework for Unified LLM Fine-tuning and Serving}},
  author = {Zhang, Yuchen and Du, Hanyue and Cao, Chun and Xu, Jingwei},
  booktitle = {Advances in Neural Information Processing Systems},
  volume = {38},
  pages = {65600--65625},
  year = {2025}
}

@inproceedings{oliaro2026flexllm,
  title = {{FlexLLM: Token-Level Co-Serving of LLM Inference and Finetuning with SLO Guarantees}},
  author = {Gabriele Oliaro and Xupeng Miao and Xinhao Cheng and Vineeth Kada and Mengdi Wu and Ruohan Gao and Yingyi Huang and Remi Delacourt and April Yang and Yingcheng Wang and Colin Unger and Zhihao Jia},
  booktitle = {23rd USENIX Symposium on Networked Systems Design and Implementation},
  pages = {1359--1379},
  year = {2026}
}

@inproceedings{qiao2024conserve,
  title = {{ConServe: Fine-Grained GPU Harvesting for LLM Online and Offline Co-Serving}},
  author = {Qiao, Yifan and Yu, Shan and Anzai, Shu and Ma, Haoran and Yang, Shuo and Wang, Yang and Kim, Miryung and Wu, Yongji and Zhou, Yang and Xing, Jiarong and Gonzalez, Joseph E. and Stoica, Ion and Xu, Harry},
  booktitle = {Proceedings of the 43rd International Conference on Machine Learning},
  series = {Proceedings of Machine Learning Research},
  volume = {306},
  pages = {101429--101443},
  year = {2026}
}

@misc{chen2026deltaserve,
  title = {{DeltaServe: Host-Agnostic Co-Serving of Inference and Fine-Tuning for LLMs}},
  author = {Chen, Jiaxuan and She, Jianshu and Yuan, Ye and Ghosh, Rajat and Gupta, Karan and Ho, Qirong and Liu, Xue and Balmau, Oana},
  year = {2026},
  note = {arXiv preprint arXiv:2607.28848}
}

@inproceedings{yu2022orca,
  title = {{Orca: A Distributed Serving System for Transformer-Based Generative Models}},
  author = {Gyeong-In Yu and Joo Seong Jeong and Geon-Woo Kim and Soojeong Kim and Byung-Gon Chun},
  booktitle = {16th USENIX Symposium on Operating Systems Design and Implementation},
  pages = {521--538},
  year = {2022}
}

@inproceedings{kwon2023efficient,
  title = {{Efficient Memory Management for Large Language Model Serving with PagedAttention}},
  author = {Kwon, Woosuk and Li, Zhuohan and Zhuang, Siyuan and Sheng, Ying and Zheng, Lianmin and Yu, Cody Hao and Gonzalez, Joseph and Zhang, Hao and Stoica, Ion},
  booktitle = {Proceedings of the 29th Symposium on Operating Systems Principles},
  pages = {611--626},
  year = {2023}
}

@inproceedings{agrawal2024taming,
  title = {{Taming Throughput-Latency Tradeoff in LLM Inference with Sarathi-Serve}},
  author = {Amey Agrawal and Nitin Kedia and Ashish Panwar and Jayashree Mohan and Nipun Kwatra and Bhargav Gulavani and Alexey Tumanov and Ramachandran Ramjee},
  booktitle = {18th USENIX Symposium on Operating Systems Design and Implementation},
  pages = {117--134},
  year = {2024}
}

@inproceedings{chen2024punica,
  title = {{Punica: Multi-Tenant LoRA Serving}},
  author = {Chen, Lequn and Ye, Zihao and Wu, Yongji and Zhuo, Danyang and Ceze, Luis and Krishnamurthy, Arvind},
  booktitle = {Proceedings of Machine Learning and Systems},
  volume = {6},
  pages = {1--13},
  year = {2024}
}

@inproceedings{sheng2024slora,
  title = {{SLoRA: Scalable Serving of Thousands of LoRA Adapters}},
  author = {Sheng, Ying and Cao, Shiyi and Li, Dacheng and Hooper, Coleman and Lee, Nicholas and Yang, Shuo and Chou, Christopher and Zhu, Banghua and Zheng, Lianmin and Keutzer, Kurt and Gonzalez, Joseph E. and Stoica, Ion},
  booktitle = {Proceedings of Machine Learning and Systems},
  volume = {6},
  pages = {296--311},
  year = {2024}
}

@inproceedings{wu2024dlora,
  title = {{dLoRA: Dynamically Orchestrating Requests and Adapters for LoRA LLM Serving}},
  author = {Bingyang Wu and Ruidong Zhu and Zili Zhang and Peng Sun and Xuanzhe Liu and Xin Jin},
  booktitle = {18th USENIX Symposium on Operating Systems Design and Implementation},
  pages = {911--927},
  year = {2024}
}

@inproceedings{xue2026muxtune,
  title = {{MuxTune: Efficient Multi-Task LLM Fine-Tuning in Multi-Tenant Datacenters via Spatial-Temporal Backbone Multiplexing}},
  author = {Chunyu Xue and Yi Pan and Weihao Cui and Quan Chen and Shulai Zhang and Bingsheng He and Minyi Guo},
  booktitle = {23rd USENIX Symposium on Networked Systems Design and Implementation},
  pages = {1533--1552},
  year = {2026}
}

@inproceedings{strati2024orion,
  title = {{Orion: Interference-aware, Fine-grained GPU Sharing for ML Applications}},
  author = {Strati, Foteini and Ma, Xianzhe and Klimovic, Ana},
  booktitle = {Proceedings of the Nineteenth European Conference on Computer Systems},
  pages = {1075--1092},
  year = {2024}
}

@inproceedings{zhang2025improving,
  title = {{Improving GPU Sharing Performance through Adaptive Bubbleless Spatial-Temporal Sharing}},
  author = {Zhang, Shulai and Chen, Quan and Cui, Weihao and Zhao, Han and Xue, Chunyu and Zheng, Zhen and Lin, Wei and Guo, Minyi},
  booktitle = {Proceedings of the Twentieth European Conference on Computer Systems},
  pages = {573--588},
  year = {2025}
}

@inproceedings{wang2025colocating,
  title = {{Colocating ML Inference and Training with Fast GPU Memory Handover}},
  author = {Jiali Wang and Yankui Wang and Mingcong Han and Rong Chen},
  booktitle = {2025 USENIX Annual Technical Conference},
  pages = {1657--1675},
  year = {2025}
}

@inproceedings{han2022microsecond,
  title = {{Microsecond-scale Preemption for Concurrent GPU-accelerated DNN Inferences}},
  author = {Mingcong Han and Hanze Zhang and Rong Chen and Haibo Chen},
  booktitle = {16th USENIX Symposium on Operating Systems Design and Implementation},
  pages = {539--558},
  year = {2022}
}

@inproceedings{ng2023paella,
  title = {{Paella: Low-latency Model Serving with Software-defined GPU Scheduling}},
  author = {Ng, Kelvin K. W. and Demoulin, Henri Maxime and Liu, Vincent},
  booktitle = {Proceedings of the 29th Symposium on Operating Systems Principles},
  pages = {595--610},
  year = {2023}
}

@inproceedings{yu2020fine,
  title = {{Fine-Grained GPU Sharing Primitives for Deep Learning Applications}},
  author = {Yu, Peifeng and Chowdhury, Mosharaf},
  booktitle = {Proceedings of Machine Learning and Systems},
  volume = {2},
  pages = {98--111},
  year = {2020}
}

@inproceedings{zhong2024distserve,
  title = {{DistServe: Disaggregating Prefill and Decoding for Goodput-optimized Large Language Model Serving}},
  author = {Yinmin Zhong and Shengyu Liu and Junda Chen and Jianbo Hu and Yibo Zhu and Xuanzhe Liu and Xin Jin and Hao Zhang},
  booktitle = {18th USENIX Symposium on Operating Systems Design and Implementation},
  pages = {193--210},
  year = {2024}
}

@inproceedings{sun2024llumnix,
  title = {{Llumnix: Dynamic Scheduling for Large Language Model Serving}},
  author = {Biao Sun and Ziming Huang and Hanyu Zhao and Wencong Xiao and Xinyi Zhang and Yong Li and Wei Lin},
  booktitle = {18th USENIX Symposium on Operating Systems Design and Implementation},
  pages = {173--191},
  year = {2024}
}

@inproceedings{wu2026fastserve,
  title = {{FastServe: Iteration-Level Preemptive Scheduling for Large Language Model Inference}},
  author = {Bingyang Wu and Yinmin Zhong and Zili Zhang and Shengyu Liu and Fangyue Liu and Yuanhang Sun and Gang Huang and Xuanzhe Liu and Xin Jin},
  booktitle = {23rd USENIX Symposium on Networked Systems Design and Implementation},
  pages = {57--74},
  year = {2026}
}

@inproceedings{lee2026tokens,
  title = {{From Tokens to Layers: Redefining Stall-Free Scheduling for MoE Serving with Layered Prefill}},
  author = {Lee, Gunjun and Kim, Jiwon and Park, Jaiyoung and Lee, Younjoo and Ahn, Jung Ho},
  booktitle = {Proceedings of Machine Learning and Systems},
  volume = {8},
  pages = {398--421},
  year = {2026}
}

@misc{vicuna2023,
  title = {{Vicuna: An Open-Source Chatbot Impressing GPT-4 with 90\%* ChatGPT Quality}},
  author = {Chiang, Wei-Lin and Li, Zhuohan and Lin, Zi and Sheng, Ying and Wu, Zhanghao and Zhang, Hao and Zheng, Lianmin and Zhuang, Siyuan and Zhuang, Yonghao and Gonzalez, Joseph E. and Stoica, Ion and Xing, Eric P.},
  year = {2023}
}

@inproceedings{bai2020pipeswitch,
  title = {{PipeSwitch: Fast Pipelined Context Switching for Deep Learning Applications}},
  author = {Zhihao Bai and Zhen Zhang and Yibo Zhu and Xin Jin},
  booktitle = {14th USENIX Symposium on Operating Systems Design and Implementation},
  pages = {499--514},
  year = {2020}
}

@manual{nvidia_cutlass_efficient_gemm,
  title = {{Efficient GEMM in CUDA}},
  author = {{NVIDIA Corporation}},
  year = {2026},
  note = {NVIDIA CUTLASS Documentation. Accessed October 26, 2026}
}

@inproceedings{bauer2012legion,
  title = {{Legion: Expressing locality and independence with logical regions}},
  author = {Bauer, Michael and Treichler, Sean and Slaughter, Elliott and Aiken, Alex},
  booktitle = {2012 International Conference for High Performance Computing, Networking, Storage and Analysis},
  pages = {1--11},
  year = {2012}
}

@article{tassiulas1992stability,
  title = {{Stability properties of constrained queueing systems and scheduling policies for maximum throughput in multihop radio networks}},
  author = {Tassiulas, L. and Ephremides, A.},
  journal = {IEEE Transactions on Automatic Control},
  volume = {37},
  number = {12},
  pages = {1936--1948},
  year = {1992}
}

@inproceedings{huang2019gpipe,
  title = {{GPipe: Efficient Training of Giant Neural Networks using Pipeline Parallelism}},
  author = {Huang, Yanping and Cheng, Youlong and Bapna, Ankur and Firat, Orhan and Chen, Mia Xu and Chen, Dehao and Lee, HyoukJoong and Ngiam, Jiquan and Le, Quoc V. and Wu, Yonghui and Chen, Zhifeng},
  booktitle = {Advances in Neural Information Processing Systems},
  volume = {32},
  year = {2019}
}

@inproceedings{narayanan2019pipedream,
  title = {{PipeDream: generalized pipeline parallelism for DNN training}},
  author = {Narayanan, Deepak and Harlap, Aaron and Phanishayee, Amar and Seshadri, Vivek and Devanur, Nikhil R. and Ganger, Gregory R. and Gibbons, Phillip B. and Zaharia, Matei},
  booktitle = {Proceedings of the 27th ACM Symposium on Operating Systems Principles},
  pages = {1--15},
  year = {2019}
}

@misc{chen2016training,
  title = {{Training Deep Nets with Sublinear Memory Cost}},
  author = {Chen, Tianqi and Xu, Bing and Zhang, Chiyuan and Guestrin, Carlos},
  year = {2016},
  note = {arXiv preprint arXiv:1604.06174}
}

@inproceedings{ren2021zerooffload,
  title = {{ZeRO-Offload: Democratizing Billion-Scale Model Training}},
  author = {Jie Ren and Samyam Rajbhandari and Reza Yazdani Aminabadi and Olatunji Ruwase and Shuangyan Yang and Minjia Zhang and Dong Li and Yuxiong He},
  booktitle = {2021 USENIX Annual Technical Conference},
  pages = {551--564},
  year = {2021}
}

@inproceedings{rhu2016vdnn,
  title = {{vDNN: Virtualized deep neural networks for scalable, memory-efficient neural network design}},
  author = {Rhu, Minsoo and Gimelshein, Natalia and Clemons, Jason and Zulfiqar, Arslan and Keckler, Stephen W.},
  booktitle = {2016 49th Annual IEEE/ACM International Symposium on Microarchitecture},
  pages = {1--13},
  year = {2016}
}

@inproceedings{osama2023streamk,
  title = {{Stream-K: Work-Centric Parallel Decomposition for Dense Matrix-Matrix Multiplication on the GPU}},
  author = {Osama, Muhammad and Merrill, Duane and Cecka, Cris and Garland, Michael and Owens, John D.},
  booktitle = {Proceedings of the 28th ACM SIGPLAN Annual Symposium on Principles and Practice of Parallel Programming},
  pages = {429--431},
  year = {2023}
}
\bibliographystyle{mlsys2025}



\end{document}